# DICTDIS: Dictionary Constrained Disambiguation for Improved NMT

**Ayush Maheshwari, Preethi Jyothi, Ganesh Ramakrishnan**
Indian Institute of Technology Bombay, India
`{ayusham, pjyothi, ganesh}@cse.iitb.ac.in`

## Abstract

Domain-specific neural machine translation (NMT) systems (*e.g.*, in educational applications) are socially significant with the potential to help make information accessible to a diverse set of users in multilingual societies. Such NMT systems should be lexically constrained and draw from domain-specific dictionaries. Dictionaries could present multiple candidate translations for a source word/phrase due to the polysemous nature of words. The onus is then on the NMT model to choose the contextually most appropriate candidate. Prior work has largely ignored this problem and focused on the single candidate constraint setting wherein the target word or phrase is replaced by a single constraint. In this work, we present DICTDIS, a lexically constrained NMT system that disambiguates between multiple candidate translations derived from dictionaries. We achieve this by augmenting training data with multiple dictionary candidates to actively encourage disambiguation during training by implicitly aligning multiple candidate constraints. We demonstrate the utility of DICTDIS via extensive experiments on English-Hindi, English-German, and English-French datasets across a variety of domains including regulatory, finance, engineering, health and standard benchmark test datasets. In comparison with existing approaches for lexically constrained and unconstrained NMT, we demonstrate superior performance for the copy constraint and disambiguation-related measures on all domains, while also obtaining improved fluency of up to 2-3 BLEU points on some domains. We also release our test set consisting of 4K English-Hindi sentences in multiple domains. The source code is present at https://github.com/ayushbits/dictdis-multi.

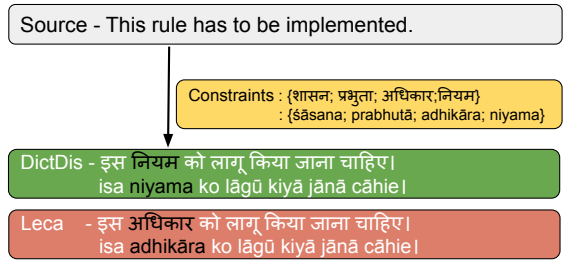

Figure 1: Example from our test set where DICTDIS chooses a contextually appropriate phrase from the multiple constraints while the constrained model Leca (Chen et al., 2021b) picks a random phrase. In addition to presenting text in the Devanagari script, transliterations are also presented in IAST format for ease of reading. In all our experiments, we present phrasal forms in Devanagari only (and not IAST).

## 1 Introduction

Neural machine translation (NMT) systems have seen great success in achieving state-of-the-art translations across several language pairs (Barrault et al., 2018). However, the default NMT pipeline does not guarantee the inclusion of specific terms in the translation output which is extremely crucial in domain-specific scenarios such as translation of technical content. While adding domain-specific terms has been relatively easier in phrase-based statistical MT, such an intervention poses a challenge in NMT owing to the difficulty of directly manipulating output representations from the decoder (Susanto et al., 2020). Alternatively, domain-specific NMT systems have been proposed to generate domain-aware translations by fine-tuning generic NMT models on domain-specific parallel text. However, such fine-tuning would require curating translation pairs for each domain, entailing significant human effort and increasing the cost of maintaining models trained separately for each domain.

Therefore, the MT output must adhere to the source domain by adopting domain-specific terminology, thus reducing and perhaps even guiding the translation post-editing effort. This is achieved in NMT via lexically constrained tech-niques that incorporate pre-specified words and phrases in the NMT output (Hokamp and Liu, 2017; Dinu et al., 2019; Chen et al., 2021b). In addition to the source sentence, word or phrasal constraints in the target language are provided as input. The constraints could be derived either from (i) in-domain source-target dictionaries or (ii) user-provided source-target constraints during interactive machine translation. Often, such constraints could encode multiple potential translations for a given source phrase. For example, the word 'speed' can be translated into 4 different Hindi phrases *teja, gati, raphtār, cāla* in the physics domain. However, existing constrained translation approaches do not accommodate such ambiguity in the constraints.

In this work, we propose a lexically constrained disambiguation framework, DICTDIS, wherein we train the NMT model to choose the most appropriate word or phrase, from among multiple constraints that are provided for a given source word or phrase. In Figure 1, we present the output of our model in contrast to another lexically constrained approach (Leca (Chen et al., 2021b)) for a sample test instance.

DICTDIS is a *copy-and-disambiguation* method that accepts a source sentence and multiple source-target constraints as input and chooses an appropriate phrase from multiple candidate translations depending on the context. During training, we sample constraints from domain-agnostic dictionaries, in which, each source phrase could have multiple candidate translations. Toward this, we create a parallel corpus by appending each source sentence with multiple candidate translations delimited by appropriate separation symbols (*c.f.*, Figure 2). The model is trained in a *soft* manner such that no constraint is forced to appear in the predicted sentence — the constraint injected should (i) be contextually relevant and (ii) not impact the fluency of the sentence. During inference, we show how DICTDIS effectively disambiguates among multiple candidate translations in domain-specific dictionaries specified at test-time (that were not available to the model during training). Our main contributions are as follows:

1) Given a source sentence and domain-specific dictionary constraints consisting of multiple source-target phrases, our proposed approach DICTDIS, can either pick the most relevant constraint or abstain from picking any constraint to least disrupt the fluency of the translation (*c.f.* Section 3.2).

2) DICTDIS is trained to do 'soft' disambiguation of phrases in the dictionary by automatically augmenting a domain agnostic corpus using synthetic domain-agnostic constraints (*c.f.* Section 3.3). Inspired by controlled text generation approaches (Pascual et al., 2021; Dathathri et al., 2019), we also introduce a user-controllable parameter in the decoding layer that can further improve the copy rate of constraints and provides a way of trading off the copy rate of constraints for fluency.

3) We present an extensive evaluation of DICTDIS on several datasets that we manually curated across finance, engineering, and regulatory domains as well as standard benchmarks (*c.f.* Section 4.2).

4) We compare against two state-of-the-art constrained approaches, *viz.*, VDBA (Hu et al., 2019), Leca (Chen et al., 2021b) as well as an unconstrained baseline. We report results using the standard BLEU (Papineni et al., 2002) metric, and a semantic matching-based metric COMET (Rei et al., 2020). Further, we estimate constraint copying rate using copy success rate (CSR) and also analyze CSR for different polysemous degrees of constraints. We observe improvements in both CSR scores coupled with fluency improvement of up to 4 BLEU score points, unlike other baselines which forgo fluency while improving CSR (*c.f.* Section 5).

## 2 Related Work

Lexically constrained NMT approaches can be broadly divided into two categories: *hard constraints* and *soft constraints*. Hard-alignment methods copy lexical constraints in the exact same form and might therefore result in the ingestion of incorrect morphological forms (Hokamp and Liu, 2017; Post and Vilar, 2018). Moreover, it often results in lower BLEU scores on out-of-domain test sets (*c.f.* Section 5). In the *hard constraints* approaches, all constraints are guaranteed to appear in the output sentence. Chatterjee et al. (2022); Chen et al. (2021a) propose a mix of alignment and soft-constraint decoding by formulating the loss function as the joint probability of token probability from the decoder and alignment probability induced from intermediate transformer layers.

*Soft constraint* approaches modify the NMT training algorithm to bias token prediction probabilities without forcing constraints to appear in the output. Ri et al. (2021); Michon et al. (2020) replace the source word with placeholders, such as a named entity type or part-of-speech tag of the tar-get constraint. Similarly, Song et al. (2019); Dinu et al. (2019) employ a bilingual dictionary to build a code-switching corpus by replacing each source phrase with the corresponding target constraint during training. However, this often interferes with the meaning of the source words, leading to poorer translation quality. (Baek et al., 2023; Chen et al., 2021b) propose a constraint-aware approach that bundles the source sentence with a single constraint using a separator token. The training data is constructed by aligning source and target sentences and randomly creating a constraint during training using the aligned pairs. D-LCNMT (Zhang et al., 2023) uses a 2-stage approach where the disambiguation module is separate from the ingestion module. It assumes that source lexicon has an additional context and attempts to align target candidates with the source lexicon. With this additional information, it uses contrastive learning to align the representation using source constraint context embedding and constraint embeddings. Then, it ingests the identified constraints using hard-constraints approach. In contrast, our approach is an end-to-end copy and disambiguation approach. Secondly, we do not assume source lexicon context is available which is an additional annotation that requires significant effort for each domain.

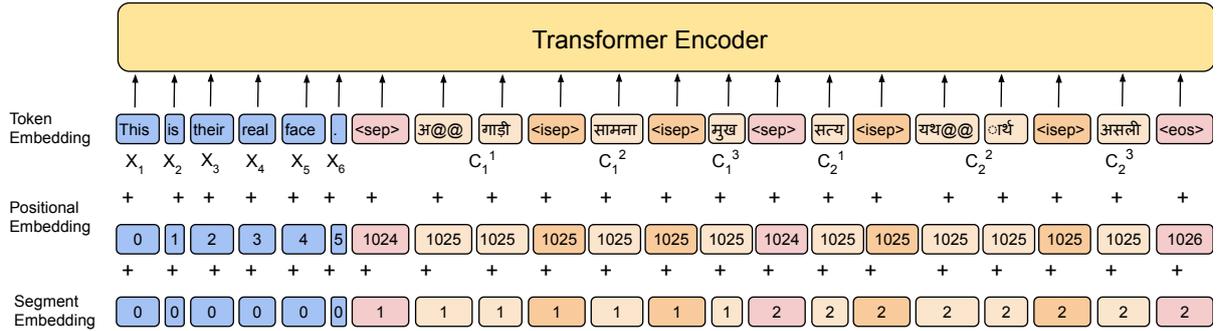

Figure 2: The embedding layer for the transformer-based encoder. Each source sentence is appended with the list of target constraints and separated by the special symbol '<sep>'. The target constraints are internally separated using the '<isep>' symbol to distinguish candidate constraints for the same source phrase.

## 3 Approach

### 3.1 Problem Statement

Let each source sentence $X = (x_1, x_2, \ldots, x_S)$ of length $S$ be associated with a constraint set $C$ derived from a dictionary. A word (or phrase) is often linked with a single constraint (Chen et al., 2021b). However, in general, a source constraint can contain one or more target constraints. Each source word or phrase can be mapped to multiple constraints in $C$. For example, $(x_i, C_i^1, \ldots, C_i^k)$ denotes $k$ possible dictionary translations associated with the word $x_i$. Let the corresponding translated sentence $Y = (y_1, y_2, \ldots y_T)$ be of length $T$. Conventional NMT systems are trained on the source and target sentences, unaware of constraints $C$. Our lexically constrained NMT system is trained with each source and target sentence, in conjunction with the possible set of constraints on the source sentence $X$.

Given a triplet of the source sentence, constraints, and target sentence $(X, C, Y)$, our objective is to choose constraint words or phrases from $C$ and potentially include them in the target sentence $Y$. (As depicted in Figure 3, the model may choose not to include any of the candidate constraints so that the flow or fluency of the predicted sentence is not adversely affected.)

### 3.2 DICTDIS: Dictionary Disambiguation in conjunction with NMT

The source sentence is appended with $C$ constraints, where each inter-phrase constraint $C_i$ is separated by a symbol '<sep>'. In turn, candidate translations $C_i^j$ for this $i^{th}$ intra-phrase constraint are separated by another symbol '<isep>'. $\hat{X} = [X, \text{<sep>}, C_1^1, \text{<isep>}, C_1^2, \text{<sep>}, C_2^1, \ldots, C_n, \text{<eos>}]$ where <eos> is the end of sentence token (c.f. Figure 2).

In our lexically constrained NMT approach, the model is trained by maximizing a log-likelihood objective similar to NMT, viz.,

$$p(Y|\hat{X};\theta) = \prod_{t=1}^{T+1} p(y_t|y_{0:t-1}, x_{1:S}, C; \theta) \quad (1)$$

Unlike Hokamp and Liu (2017); Post and Vilar

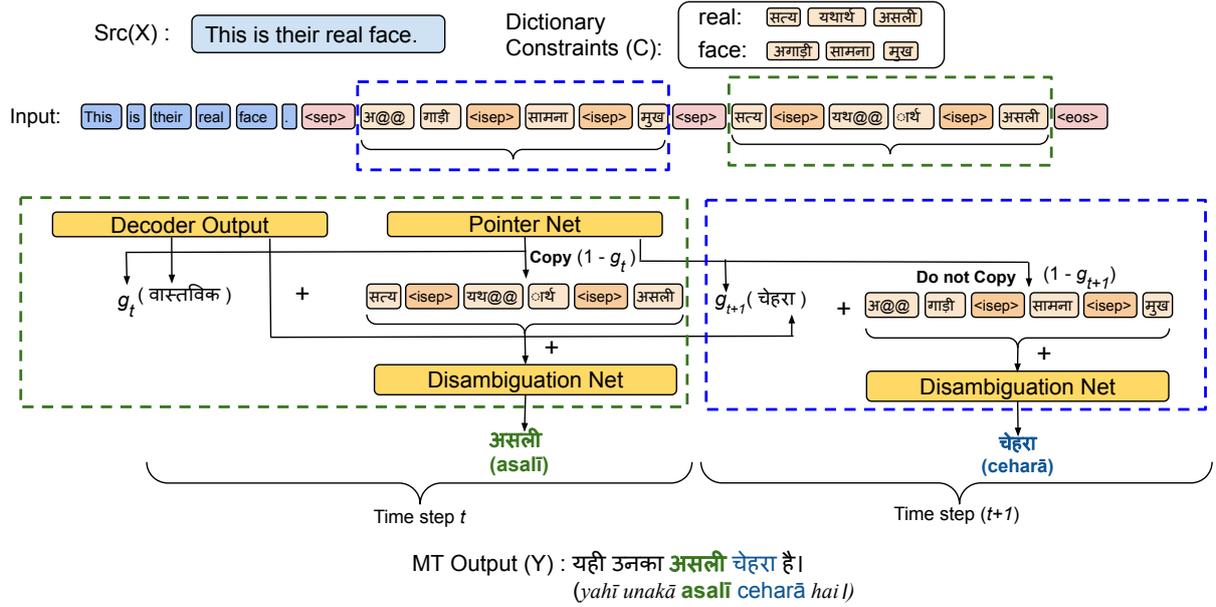

Figure 3: DICTDIS decoding mechanism for English to Hindi translation. Gating value $g_t$ regulates the final token probability which determines the decision of whether to disambiguate or to copy.

(2018), our method is based on *soft* ingestion, in which constraints are not forced to appear in the output. The logit values of constraints are adjusted during training such that the constraints occur in the beam (within the specified size) while decoding.

### 3.2.1 Encoder

We employ the standard Transformer-based architecture for NMT (Vaswani et al., 2017) that uses self-attention networks for both encoding and decoding. The encoder is a stack of $N$ identical layers, each of which contains two sub-layers. Each layer consists of a multi-head self-attention (Self-Att) and a feed-forward neural network (FFNN). For time step $i$ in layer $j$, the hidden state $h_{i,j}$ is computed by employing self-attention over hidden states in layer $(j-1)$.

We define a segment as the sequence of constraints corresponding to the textual span between two <sep> symbols. Inspired by BERT (Devlin et al., 2018), we append our input source token embedding with the learned embedding for the segment. This is illustrated in Figure 2. Further, positional indices of the constraint tokens begin with a number that is larger than the maximum source sentence length. For a given token, its embedding is the sum of three components, *viz.*, token embedding, positional embedding, and segment embedding.

### 3.2.2 Decoder

With a structure similar to the encoder, the decoder consists of a stack of $N$ identical layers. In addition to two sub-layers, it also contains a cross-attention (CrossAtt) sub-layer to capture information from the encoder. Cross attention is computed between the last layer of the encoder hidden state $h = h_{1,n}, h_{2,n} \ldots h_{m,n}$ and the output of the self-attention sub-layer, $\tilde{s}_l$. Each sub-layer output is followed by a layer normalization step (Ba et al., 2016). The probability for the token $t$ is computed through a softmax over the target-side vocabulary, applied to the final decoder state $s_t$ as

$$P_t^{pred}(y_t|(y_{<t}, \hat{X})) = \text{softmax}(s_{t,N}, W) \quad (2)$$

where $W$ is the learnable weight matrix, $\hat{X}$ is the source sentence with target constraints and $y_1, y_2 \ldots y_t$ represent the target phrases and N is the final decoder block in the stack. Inspired by (Gulçehre et al., 2016) and (Chen et al., 2021b), we introduce a pointer network that adds logit values over the target-side constraints. Specifically, the token probability over the target vocabulary is the weighted sum of decoder probabilities from the predictive model $P_t^{\text{pred}}$ and copy probabilities $P_t^{\text{copy}}$. This strengthens copying by identifying the target constraint (provided as an input) that needs to be copied. Copy probability for time step $t$ is computed as the average multi-head attention weights of the last decoder layer. Intuitively, $P_t^{\text{copy}}$ is the attention weight for the corresponding source position $s$.

$$P_t^{\text{copy}} = \text{avg}_k \left( \text{CrossAtt}(\tilde{s}_{t,N}, h^k, h^k) \right) \quad (3)$$

$\text{avg}_k$ denotes averaging attention weights obtained via cross-attention across all the attention heads (indexed by $k$) and $\tilde{s}_{t,N}$ is the output of $N-th$ self-attention layer for token $t$.

We further add a disambiguation network to disambiguate between multiple inter-phrase constraints. This network learns to differentiate between multiple senses of a constraint based on the sentence context. We define the disambiguation score for the $j^{\text{th}}$ inter-phrase component of the $i^{\text{th}}$ constraint, $P_{ij}^{dis}$ as the dot product of the attention-weighted context vector and the $j^{\text{th}}$ inter-phrase component's contextual embedding (normalized overall $j$ inter-phrase constraints). We anticipate that constraint-component embedding combined with source context and cross-attention can distinguish between multiple inter-phrase constraints.

$$\text{score}_{ij}^{dis} = c_t . e_i^j = \sum_{s=1}^{\hat{S}} \alpha_{t,s} h_s e_i^j$$
$$P_{ij}^{dis} = \text{normalise}_j(\text{score}_{ij}^{dis}) \quad (4)$$

where $c_t = \sum_{s=1}^{\hat{S}} \alpha_{t,s} h_s$ is the attention-weighted context vector and $e_i^j$ is the contextual embedding of inter-phrase constraints. Recall that $h_s$ is the encoder hidden state at position $s$ of the last layer, $\alpha_{t,s}$ is the averaged attention weight at the last decoder layer for source position $s$ at decoding time step $t$.

$P_t^{copy} + P_t^{dis}$ is a normalised probability distribution so that the weighted combination remains a probability distribution. The final distribution over the target vocabulary is defined as the weighted sum of $P_t^{pred}$ and the normalized probability sum of $P_t^{copy} + P_t^{dis}$ (refer Fig 3). :

$$p(y_t|(y_{<t}, \hat{X})) = g_t P_t^{pred} + (1 - g_t)(P_t^{copy} + P_t^{dis}) \quad (5)$$

where $g_t \in [0, 1]$ is a *learnable* gate controlling the weightage of the two probability distributions. It decides whether to keep the provided user constraint in translation or let the model predict (as can be seen in Fig. 3). We compute $g_t$ as $g_t = \text{FeedForward}(c_t, s_t)$ where $c_t$ is the weighted context vector, $s_t$ is the hidden state of the last layer of the decoder until timestep $t$, and FeedForward is a single layer FFNN.

### 3.3 Domain-agnostic Training

**Data**. We conduct experiments on English-Hindi, English-German, and English-French translation tasks. To train DICTDIS we construct constraints using a domain-agnostic source-target dictionary such that each source phrase can contain one or more target phrases. We use a bilingual dictionary to append constraints if a phrase in the source sentence matches the source side of the dictionary. We train our model on the Samanantar dataset (Ramesh et al., 2022) consisting of 8.4 million English-Hindi parallel sentences. We use a generic English-Hindi dictionary[1] containing 11.5K phrases. In Appendix B, we present the distribution of phrases (in percentage) in this domain-agnostic dictionary, with respect to the number of constraints associated with each phrase. During pre-processing, we match source-side dictionary phrases with the parallel corpus and find that 96% of the sentences contain at least one constraint pair. To avoid adversely affecting the translation performance (in terms of fluency), we leave the remaining 4% sentences as unconstrained.

In the case of English-German, we exactly follow the setup of Leca (Chen et al., 2021b) and employ the WMT16 news data as a training corpus consisting of 1.8M sentences, newstest2013 as the development set, and newstest2014 as the test set. For English-French, we use the Europarl corpus for training consisting of 2M sentences and terminology tasks from WMT 21 (Alam et al., 2021) as validation and test set. To train DICTDIS, we use En-De and En-Fr bilingual dictionary from Muse dataset (Lample et al., 2018). We use Moses tokenizer (Koehn et al., 2007) for pre-processing En-De, En-Fr, and the Indic-NLP (Kunchukuttan, 2020) for pre-processing En-Hi sentences. The tokenized sentences are then processed using BPE (Sennrich et al., 2016) with 32K merge operations for the language pairs. We detokenize the predictions before computing BLEU, COMET, and CSR. We describe implementation details in Section C in the Appendix.

## 4 Experiments

### 4.1 Evaluation

The performance of constrained machine translation is evaluated using the following three metrics, *viz.*, (1) **BLEU**: The BLEU score (Papineni et al., 2002) measures the translation quality by comparing n-grams of the predicted translation with respect to the reference translation. We use SacreBLEU (Post, 2018) to estimate the BLEU score. BLEU scores fail to robustly match paraphrases and measure semantic consistency between predicted and reference translation. This results in a low BLEU score for fluent sentences having diverse translations. Therefore, we also report our results using the (2) **COMET** (Rei et al., 2020) metric which is based on a pre-trained language model and has shown a higher correlation with human judgments. We use the pre-trained `wmt20-comet-da model` for reporting COMET scores. We report COMET scores in Table 7 in the Appendix. (3) **CSR** (Copy Success Rate) (Campolungo et al., 2022): Following previous works (Chatterjee et al., 2022; Chen et al., 2021b; Song et al., 2019), we report CSR which measures the percentage of constraints that are successfully generated in the translation. If constraints with multiple candidates are present during translation, CSR rewards the correct copying of constraints with respect to the ground truth sentences.

---
[1]Available at https://sanskritdocuments.org/hindi/dict/eng-hin_unic.html.

| Testsets | #sentences | Dictionary | Polysemous Degree (in %) | | | | |
|---|---|---|---|---|---|---|---|
| | | | 1 | 2 | 3 | 4 | 5 |
| Regulatory | 1000 | Banking | 90.4 | 9.5 | 0.1 | - | - |
| Aerospace | 1000 | Phy,Chem, Maths, Mech | 32.3 | 38.7 | 15.3 | 8.4 | 2.6 |
| Banking | 1016 | Banking | 85 | 14.8 | 0.1 | - | - |
| Flores | 1012 | Administrative | 26 | 36 | 16.2 | 15.9 | 2.9 |
| WAT2021 | 3003 | Administrative | 32.3 | 38.8 | 15.4 | 8.4 | 2.6 |
| WMT14 | 3003 | Wiktionary | 84.5 | 14.1 | 1 | - | - |
| WMT2021 | 2100 | Covid-19 | 11.5 | 34.8 | 18.5 | 24 | - |

Table 1: Test sets used in our experiments. 'Dictionary' refers to the type of domain dictionaries used for constrained translation. Polysemous degree refers to the number of candidate target constraints in the dictionary corresponding to a source constraint. The numbers represent the distribution of target constraints for the corresponding polysemous degree in the test set. Due to the paucity of space, we show constraints up to polysemy degree $\leq 5$.

### 4.2 Test Sets

In Table 1, we present a summary of manually curated domain-specific parallel corpora in different domains to evaluate the performance of DICTDIS with domain-specific dictionary constraints. To the best of our knowledge, there is no such domain-specific publicly available aligned parallel corpus for Indian languages. Two of our corpora (*viz.*, Banking and Regulatory) were curated by aligning sentences from publicly available reports. The aerospace corpus was developed by manually translating sentences from an undergraduate textbook of aerospace engineering by a team of translators and reviewers. In addition to the curated corpus, we also present comparisons with popular benchmarks *viz.*, Flores-101 (Goyal et al., 2022) and WAT 2021[2]. We use administrative dictionaries on these two datasets during constrained inference. We provide details on the annotation effort of each of our in-house developed datasets in Appendix A. For En-De, following Chatterjee et al. (2022), we use Wiktionary.975[3] as our constraint dictionary. For En-Fr, we use the Covid-19 dictionary as provided in the WMT shared task on terminologies (Alam et al., 2021).

### 4.3 Baselines

We compare against state-of-the-art constrained as well as unconstrained approaches.
1. **Leca** (Chen et al., 2021b): This is a constrained approach that accepts a single target constraint and the source sentences as input. In case of multiple constraints for a phrase, Leca chooses a random phrase for insertion.
2. **Vectorized Dynamic Beam Allocation** (VDBA) (Hu et al., 2019) extends beam search to include pre-specified lexical constraints in the generated translation. We use a constrained decoding implementation of fairseq to run the inference over an unconstrained transformer.
3. **Transformer** refers to an unconstrained base transformer trained using fairseq.
4. **DICTDIS(Uncons)** refers to the DICTDIS model (trained with dictionary constraints), however during inference, we do not provide additional constraints. This helps us in comparison with the DICTDIS and assess whether there is any degradation/improvement in performance as a result of adding constraints with the input. On the 'with constraint set' in Table 2, the objective of DICTDIS(Uncons) is to compare with the Transformer which is also devoid of constraints during inference. On the 'without constraint test sets', DICTDIS and DICTDIS(Uncons) have similar input, therefore, numbers are left blank for the DICTDIS(Uncons). The aim of adding

---
[2]http://lotus.kuee.kyoto-u.ac.jp/WAT/indic-multilingual/index.html
[3]https://github.com/mtresearcher/terminology_dataset/

| Model | En-Hi | | | | | | | | | | En-De | | En-Fr | |
|---|---|---|---|---|---|---|---|---|---|---|---|---|---|---|
| | Banking | | Aerospace | | Regulatory | | Flores | | WAT 2021 | | WMT 14 | | WMT 21 | |
| | w/o | with | w/o | with | w/o | with | w/o | with | w/o | with | w/o | with | w/o | with |
| Transformer | 36.7 | 33.1 | 44.7 | 42.9 | 28.5 | **26.9** | 30.9 | 32.3 | 35 | 36.2 | 34.7 | 34.5 | 35 | 33.2 |
| VDBA | - | 18 | - | 16.8 | - | 16.7 | - | 16.4 | - | 15.3 | - | 31.7 | - | 32.4 |
| Leca | 36.3 | 28.5 | 40.5 | 32.6 | 28.5 | 25.4 | 30.2 | 28 | **35.1** | 29.6 | 33.6 | 32.1 | 34.3 | 33.6 |
| DICTDIS(Uncons) | - | **34.1** | - | 43.5 | - | 26.5 | - | 32.9 | - | **37.1** | - | **35.2** | - | **34.6** |
| DICTDIS | **37** | 33.7 | **45.1** | **44** | **28.9** | 26.3 | **32.2** | **33.1** | 34.8 | 36.8 | **35.9** | 35 | **35.5** | 34.1 |

Table 2: BLEU scores of constrained NMT without (w/o) and with constraints on the test sets. DICTDIS(Uncons) refers to the unconstrained setting where the input sentence to the DICTDIS model is devoid of target constraints. VDBA has the same results as Transformer without constraint on the test sets. Underlines refers to statistically significant difference between Transformer (top row) and DICTDIS(last row) at p< 0.05. We report corresponding COMET scores in Table 7 in the Appendix.

| Model | En-Hi | | | | | En-De | En-Fr |
|---|---|---|---|---|---|---|---|
| | Banking | Aerospace | Regulatory | Flores | WAT 2021 | WMT 14 | WMT 21 |
| Transformer | 77.1 | 81.6 | 79.3 | 83.2 | 90.4 | 82.8 | 72.5 |
| Leca | **91.7** | 75.6 | **94.1** | 72.9 | 78.0 | 85.5 | 68.2 |
| DICTDIS(Uncons) | 75.1 | 79.7 | 78.7 | 83.7 | 90.4 | 83.1 | 72.3 |
| DICTDIS | 83.8 | **82.2** | 85.3 | **84.4** | **91.3** | **85.9** | **74.9** |

Table 3: Copy success rate (CSR) results on the constrained test sets. Leca (Chen et al., 2021b) achieves higher CSR on test sets (*viz.* Banking and Regulatory) having a higher percentage of single target candidate constraints. However, it suffers from low BLEU scores due to aggressive single-constraint ingestion. We omit VDBA since it has CSR greater than 99% on all the datasets.

DICTDIS(Uncons) is to a) compare with the Transformer approach where no constraints are added as input either during training and inference, and b) compare with DICTDIS to assess the impact of passing constraints during inference on the BLEU score.

## 5 Results

We present BLEU scores of DICTDIS and all baselines in Table 2. We split our test set into constrained sets by matching at least a pair of source and target constraints on both source and target sentences. An unconstrained test set refers to those sentences where constraints from a bilingual dictionary are not present in both source and target sentences. For unconstrained test sets, our method DICTDIS outperforms all the baselines. We observe that all models preserve their ability to translate unconstrained test sentences.

On the constrained test set, DICTDIS and its unconstrained variants achieve the best BLEU performance on 5 test sets. DICTDIS yields the best performance on 2 test sets and near-best scores on the other 4 test sets. For the Aerospace and Flores test set, DICTDIS achieves the best performance. On Banking, WAT 2021, and WMT 14 test sets, the un-constrained variant of DICTDIS achieves the best performance while the constrained variant trails by 0.4, 0.3, and 0.2 points respectively. On 5 out of 7 test sets, unconstrained DICTDIS performs marginally better than the constrained version. The BLEU score difference between DICTDIS(uncons) and DICTDIS(cons) is 0.2-0.5 on the test sets. This could be possibly due to minor impact on fluency due to constraint ingestion. The greedy beam search decoding approach is affected by the constraints ingested in the previous step(s) due to which marginal degradation in BLEU scores are observed. Despite achieving similar scores, DICTDIS(cons) reports better CSR that unconstrained on all the test sets in Table 3. On the other hand, DICTDIS(uncons) demonstrates the advantage of training with a constrained decoding method.

Leca performs poorly on all the constrained test sets. The difference between Leca and the best-performing method on the Banking, Aerospace, Regulatory, Flores, WAT2021, and WMT14 set is 5.6, 11.4, 1.4, 5.1, 7.2, and 3.1 BLEU scores respectively. VDBA (Hu et al., 2019) enforces constraints only considering the target tokens of the lexicons which reduce the fluency of the sentence. VDBA algorithm enforces constraints to appear in the output which is reflected in close to 99% CSR scores for the test sets but very low BLEU scores on out-of-domain test sets. Leca has far better BLEU scores than VDBA due to its training framework which focuses on learning a gating value of constraints when the decoder-encoder attention over the constraints has some significant value.

| Test Sets | Banking | | Aerospace | | | | | | Regulatory | | Flores | | | | WAT2021 | | | | | WMT14 (En-De) | | | WMT21 (En-Fr) | | | |
|---|---|---|---|---|---|---|---|---|---|---|---|---|---|---|---|---|---|---|---|---|---|---|---|---|---|---|
| Polysemy → | 1 | 2 | 1 | 2 | 3 | 4 | 5 | 6 | 1 | 2 | 1 | 2 | 3 | 4 | 1 | 2 | 3 | 4 | 5 | 1 | 2 | 3 | 1 | 2 | 3 | 4 |
| Leca | 95 | 70 | 95 | 57 | 68 | 45 | 51 | 66 | 96 | 74 | 98 | 65 | 65 | 54 | 96 | 74 | 68 | 60 | 56 | 88 | 64 | 65 | 93 | 67 | 68 | 59 |
| DictDis | 82 | 90 | 84 | 79 | 83 | 82 | 72 | 83 | 85 | 88 | 91 | 82 | 79 | 83 | 93 | 92 | 87 | 85 | 87 | 89 | 72 | 66 | 74 | 66 | 79 | 89 |

Table 4: Results for polysemy degree-wise CSR for different test sets and constrained approaches.

| Model | Banking | | Regulatory | |
|---|---|---|---|---|
| | BLEU | CSR | BLEU | CSR |
| Leca | 28.5 | 91.7 | 25.4 | 94.1 |
| DictDis | 33.7 | 82.1 | 26.3 | 85.3 |
| DictDis + $\alpha$ | 32.9 | 87.8 | 25.9 | 86.7 |

Table 5: BLEU and CSR for Banking and Regulatory dataset with constraint ingestion parameter $\alpha = 0.1$.

We perform a paired significance test between DictDis and Transformer baseline at p < 0.05 (Koehn, 2004) with bootstrap resamples as 1000. In Table 2, we highlight statistical significant scores with an underline.

### 5.1 Copy Success rate (CSR)

In Table 3, we present CSR of various baselines and DictDis. DictDis yields better CSR scores for Aerospace and WAT2021, whereas Leca has better CSR on Regulatory, Banking, and WMT14 test sets. Leca achieves better CSR scores on test sets where single target constraints form a major part of a constrained set (*c.f.*, Table 1) such as Banking, Regulatory, and WMT14. On the contrary, it performs poorly on constraints having a polysemous degree >1. DictDis constrained setting performs second best on the Banking, Regulatory, and WMT14 datasets and yields the overall best performance on other test sets.

While Leca only utilizes a copy network, DictDis takes advantage of both copy and disambiguation networks. In Table 4, we present polysemy degree-wise CSR scores for different test sets and constrained approaches. As pointed out in the preceding discussion, Leca ingests single-degree constraints aggressively; however, it is unable to pick appropriate constraints when the polysemy degree is 2 or higher. DictDis performs consistently better for polysemy degree >2 and yields an overall higher BLEU score on most datasets.

Low disambiguation performance on Regulatory and Banking can be explained by the inability of DictDis to ingest unambiguous (*i.e.*, polysemy degree 1) constraints aggressively in the predicted sentence. The Regulatory dataset includes around 90% unambiguous constraints from the banking dictionary while other engineering and medical datasets have a sizeable proportion of constraints with degree >1. In the case of Regulatory, Leca ingests unambiguous constraints aggressively, resulting in much better disambiguation performance than other approaches. On the Banking dataset, Leca is unable to disambiguate ambiguous phrases with degree 2 polysemy. On other datasets, DictDis has better disambiguation performance and superior BLEU scores.

### 5.2 Controlled Text Generation

We observe that DictDis is not aggressive in ingesting single-degree constraint in favor of fluency. Therefore, inspired by controlled text generation approaches (Pascual et al., 2021; Dathathri et al., 2019), we introduce a user-controllable parameter $\alpha$ which controls the aggressiveness of ingesting single-degree constraints. $\alpha$ is governed by the normalized cross attention on the constraints for each time step $t$. The final value of logits of constraints is increased by $\alpha \times CrossAtt_t$. In Table 5, we present results for $\alpha = 0.1$ which is the optimum value governing the tradeoff between CSR and BLEU. We observe that DictDis + $\alpha$ achieves better CSR than DictDis on these test sets without sacrificing fluency, unlike Leca. We observe that higher values of $\alpha$ result in high CSR but at the expense of lower BLEU scores. In Table 8, 9 and 10 in Appendix E, we present examples of translations with constrained approaches, *viz.*, DictDis and Leca.

## 6 Conclusion

We present a *copy-and-disambiguation* approach, *viz.*, DictDis to translate under dictionary constraints provided at run-time, that potentially include multiple target candidates for each source language phrase. We present a recipe for training DictDis on a generic parallel corpus by synthesizing constraints during training by leveraging domain-agnostic dictionaries. We present an exten-sive evaluation of DICTDIS on existing datasets as well as on several new datasets, that we manually curated from the finance and engineering domains.

## 7 Limitations

A major limitation with training the current framework is the need for bilingual synonym dictionaries to have good coverage of words/phrases in the source sentences. Although synonym dictionaries are available for high-resource languages[4], they are not readily available for low-resource languages. Secondly, our framework is limited by the maximum length of constraints. In cases where the length of constraints is greater than 1024 characters, our model chooses to ignore such constraints. Thirdly, model training time is 3x times higher than the base transformer primarily due to higher input length.

## Acknowledgements

We thank Piyush Sharma for his contributions in the initial experiments of the paper. We thank Anuja Dumada, Pranita Harpale and Atul Kumar Singh for preparing few evaluation datasets. Ayush Maheshwari did this work as part of the PhD at IIT Bombay and was supported by a fellowship Ekal Foundation during his PhD. Ganesh Ramakrishnan is grateful to the National Language Translation Mission (NLTM): Bhashini project by Government of India and IIT Bombay Institute Chair Professorship for their support and sponsorship.

---

[4]https://github.com/facebookresearch/MUSE

# Appendix

## A  Dataset

We adopted following two methods to create the parallel corpora for English-Hindi test set:

1. **Alignment:** We align parallel data available in the finance domain. Several banking organizations in India produce their work reports in both Hindi and English. We semi-automatically align around 3K sentences using our in-house developed tool (Maheshwari et al., 2023).

2. **Manual Translation**: We obtained manual translations for engineering curriculum books from English into Hindi with the help of professional translators. This was with requisite permissions from the corresponding publishers and authors.

i) **Regulatory**: These are manually translated sentence pairs of the annual report of a central bank. We extract the sentences using OCR and automatically align sentences using multilingual sentence embeddings. The aligned sentences are then manually reviewed for alignment and OCR errors (as adopted in Maheshwari et al. (2022)). The dictionary here primarily belongs to the finance domain. We used the Universal Sentence Encoder Model[5] to compute sentence embedding for both English and Hindi sentences. Then, we compute the dot product and pair the closest matching sentence above the heuristically defined threshold. Secondly, we ask a human annotator to verify the automatically aligned sentences. We observe around 95% accuracy (with a drop of around 15% sentences due to thresholding).

ii) **Aerospace**: This is an undergraduate aerospace engineering book. Since aerospace is a multi-disciplinary field, we choose dictionaries from physics, chemistry, mechanical, and math domains to use with DICTDIS. Any single dictionary may not provide sufficient enough coverage over domain-specific terms.

iii) **Banking**: These are manually translated sentence pairs of the annual report of a banking organization. We adopt the same method as in the regulatory dataset to derive the test set. We recruited expert linguistic translators who are domain experts and worked as part of at least 3 book translation project. We provided source phrases in Microsoft Excel and asked translators to check the corresponding translations. All the translators were duly compensated as per industry standards.

---

[5] https://tfhub.dev/google/universal-sentence-encoder/4

## B Training Data

We use a generic English-Hindi synonym dictionary containing 11.5K phrases. In Table 6, we present the distribution of phrases (in percentage) in this domain-agnostic dictionary across different numbers of constraints associated with a phrase. During pre-processing, we match source-side dictionary phrases with the parallel corpus and find that 96% of the sentences contain at least one constraint pair. To avoid adversely affecting the translation performance (in terms of fluency), we leave the remaining 4% sentences as unconstrained.

| #constraints | % of phrases |
|---|---|
| 1 | 15.6 |
| 2 | 27.6 |
| 3 | 29.6 |
| 4 | 19.7 |
| 5 | 7.0 |
| 6 | 0.85 |

Table 6: Percentage of phrases having different numbers of constraints in the constraint dictionary during training.

## C Implementation Details

We implement DICTDIS using fairseq toolkit (Ott et al., 2019) v0.12[6] over base Transformer model (Vaswani et al., 2017) for all our experiments. The results are reported for the single run. We use the Byte-Pair Encoder (BPE) tokenizer () with a maximum vocabulary size of 32000 for all our experiments. We set the maximum token length (including all inter-phrase constraints and intra-phrase constraints) to 2048. The optimizer employed is Adam (Kingma and Ba, 2014) with label smoothing of 0.1, the learning rate is set to 5e-4 with 4000 warm-up steps, the probability dropout is set to 0.3, maximum token length in a batch to 4096 and maximum number of updates to 200,000. The beam size for all experiments and baselines is set to 5. To maintain consistency in the length of constraints in the batch, we pad tokens to make their lengths equal. Training takes approximately 3 hours for 1 epoch on four Nvidia A6000 GPUs in a distributed setup.

---
[6]https://github.com/facebookresearch/fairseq/

## D COMET scores on Test sets

In Table 7, we present mean COMET scores for the constrained test sentences i.e. test sentences containing both source and target-side constraints. COMET (Rei et al., 2020) metric is based on a pre-trained language model that accepts source, reference, and predicted sentence as input. COMET has shown a higher correlation with human judgments and its ability to identify the highest-performing MT systems. We use pre-trained `wmt20-comet-da model` for calculating COMET scores. We observe that DICTDIS achieves the highest rank among various baselines on 5 out of 6 test sets. En-De language pair has a small/negative score for a few baselines which could possibly be due to the different scores scales. However, DICTDIS ranks higher for the majority of test sets.

## E Examples for qualitative illustration

In Table 8, 9 and 10, we present example translations from Flores, Aerospace dataset, and Regulatory dataset respectively. We present outputs with constrained approaches, *viz.*, DICTDIS and Leca, and the unconstrained Transformer. For the specified input sentence in English and corresponding dictionary constraints, our approach incorporates constraints without adversely affecting the fluency of the sentence. In the examples, DICTDIS incorporates constraints specified during runtime into the translation and produces better sentence translations. Since our method is trained with multiple polysemy constraints, it learns to choose (or drop) appropriate constraints.

|  | Banking | Aerospace | Regulatory | Flores | WAT2021 | WMT14 |
|---|---|---|---|---|---|---|
| VDBA | 71.6 | 65.6 | 62.3 | 51.9 | 94.2 | -3.7 |
| Leca | 64.4 | 48.9 | 60.9 | 51.4 | 78.5 | -13.1 |
| Transformer | 71.4 | 66.8 | **71.4** | 62 | 94.6 | 2.9 |
| DICTDIS | **74.5** | **77.3** | 65.7 | **72.8** | **97.5** | **8.2** |

Table 7: COMET scores for the constrained test sets. Note that COMET scores are representative of ranking relative to direct assessment as measured by the pre-trained model instead of the translation quality.

| Model | Translations |
|---|---|
| Input | From 11 km onwards , the **temperature** is **constant** up to an altitude of 20 km . |
| Constraints | ('constant, 'niyata, **sthira**, acala, aparivartī sthirāṃka, ekasamāna,satata'),('temperature', '**tāpamāna**, tāpakrama) |
| Leca | 11 kilomīṭara se lekara 20 kilomīṭara taka tuṃgatā kā <u>tāpakrama aparivartī sthirāṃka</u> meṃ banā huā hai |
| DictDis | 11 kimī se, **tāpamāna** 20 kimī kī ūṃcāī taka **sthira** hai |
| Reference | 11 kimī se **tāpamāna** 20 kimī kī ūṃcāī taka **sthira** hai |
| Input | Israels **era** of **rule** by judges ended with the **coronation** of its **first** king, Saul . |
| Constraints | ('rule', '**śāsana**, prabhutā,adhikāra,niyama')('first', 'prathama, **pahilā**, sarva prathama, mukhya') ('era', 'samvata, varṣa, sana, **kāla**')('coronation', 'rājatilaka,ṭīkā, rājagaddī') |
| Leca | <u>ṭīkā karane ke varṣa</u> meṃ, isrāela ke <u>prathama</u> rājā, śāula ke **rājyābhiṣeka** ke sātha nyāya karane kā <u>adhikāra</u> samāpta huā |
| DICTDIS | nyāyiyoṃ dvārā isrāeltranslationsa kā **śāsana kāla** apane **pahale** rājā, śāula ke **rājyābhiṣeka** ke sātha samāpta huā |
| Reference | nyāyiyoṃ dvārā **śāsana** kie jāne kā isrāela kā **kāla** usake **pahale** rājā, śāula ke **rājyābhiṣeka** ke sātha samāpta ho gayā |

Table 8: Examples from the test set using DICTDIS and Leca (constrained) models. Here, the correctly ingested constraints are highlighted in **bold** and incorrectly ingested constraints are highlighted as <u>underlined</u>. Sentences have been written in the IAST format for readability. DICTDIS is generally more effective in ingesting the constraints appropriately than Leca while maintaining fluency.

| Model | Sentences |
|---|---|
| Input | Stability and Turn Down Ratio The operation of the combusting device must be stable over the operating range. |
| Constraints | ('ratio', '**anapātu**,anupāta'), (operation, 'saṃkriyā,**pracālana**'), (operating,'saṃkāraka,**pracālana**'), ('down', 'adha,nīce'),('range', 'parāsa parisara mālā,śreṇī, parāsa,parisara,mālā,śreṇī'), ('stability', '**sthāyitva**') |
| DictDis | <u>sthiratā</u> aura ṭarna **ḍāuna anupāta** kambasṭiṃga yukti kā **pracālana pracālana** sīmā para sthira honā cāhie | |
| Transformer | kambasṭiṃga <u>divāisa</u> kā **saṃcālana** <u>auparetiṃga remja</u> para sthira honā cāhie | |
| Leca | **sthāyitva** aura ṭarna emda **ḍāuna anupāta** kambasṭiṃga yukti kī <u>saṃkriyā auparetiṃga śreṇī</u> para anakāraka honī cāhie | |
| Reference | **sthāyitva** aura ṭarna **ḍāuna anupāta**: kambasṭiṃga yukti kā pracālana pracālana sīmā para sthira honā cāhie | |
| Input | The precise mechanics of such a transformation has not been satisfactorily worked out yet. |
| Constraints | ('transformation', '**rūpāṃtaraṇa**')('precise', 'pratatha,pariśuddha') ('mechanics', 'balavijñāna,yaṃtravijñāna,**yāṃtrikī**')('not', 'nadta') |
| DictDis | isa taraha ke <u>parivartana</u> kī satīka **yāṃtrikī** abhī taka saṃtoṣajanaka dhaṃga se nahīṃ banāī gaī hai | |
| Transformer | isatranslations prakāra ke **rūpāntaraṇa** kī yathārtha <u>kriyāvidhi</u> kā abhī taka santoṣajanaka rūpa se patā nahīṃ calā hai | |
| Leca | isa taraha ke **rūpāṃtaraṇa** kā pravijñāna abhī taka saṃtoṣajanaka dhaṃga se <u>nadta-nadta</u> nahīṃ huā hai | |
| Reference | isa taraha ke **rūpāṃtaraṇa** kī satīka **yāṃtrikī** para abhī saṃtoṣajanaka dhaṃga se kāma nahīṃ kiyā gayā hai | |

Table 9: Constrained translation examples from the Aerospace test set using DICTDIS, Transformer (unconstrained) and Leca (constrained) models. Here, the correctly ingested constraints are highlighted in **bold**, and incorrectly ingested constraints are highlighted as <u>underlined</u>. Sentences have been written in the IAST format for readability.

| Model | Translations |
|---|---|
| Input | Forward exchange contracts are valued half-yearly, and net loss, if any, is provided for in the Exchange Equalisation Account ( EEA ). |
| Constraints | ('Net Loss', 'nivala hāni') , ('Forward exchange contracts', 'vinimaya samvidāem') |
| Transformer | pharavara eksacemja anubamdhom kā mūlya chamāhī hotā hai aura <u>śuddha hāni</u>, yadi koī ho, vinimaya samakārī khāte (EEA) mempradāna kī jātī hai |
| DictDis | vāyadā **vinimaya saṃvidāoṃ** kā mūlyāṃkana arddha vārṣika hotā hai aura **nivala hāni**, yadi koī ho, eksacemja ikvīlāijeśana akāuṃṭa (īe) mem pradāna kī jātī hai |
| Reference | vāyadā vinimaya saṃvidāoṃ kā mūlya chamāhī ādhāra para nikālā jātā hai aura nivala hāni, yadi koī ho, kā vinimaya samakaraṇa khāte meṃ prāvadhāna kiyā jātā hai |
| Input | The Union Budget aims to keep the economy on the path of fiscal consolidation. |
| Constraints | ('fiscal consolidation', 'rājakoṣīya sudṛḍhaīkaraṇa') |
| Transformer | kendrīya bajaṭa kā uddeśya arthavyavasthā ko rājakoṣīya <u>samekana</u> ke mārga para banāe rakhanā hai |
| DictDis | keṃdrīya bajaṭa kā uddeśya arthavyavasthā ko **rājakoṣīya sudṛḍhaīkaraṇa** ke patha para banāe rakhanā hai |
| Reference | saṃghīya bajaṭa kā yaha lakṣya hotā hai ki arthavyavasthā ko rājakoṣīya samekana ke patha para rakhā jāe |
| Input | Notable gains in containment of key deficit indicators in 2013-14 are apparent in the provisional accounts ( PA ). |
| Constraints | ('key deficit indicators', 'mukhya ghāṭā saṃketaka') |
| Transformer | 2013-14 meṃ <u>pramukha</u> ghāṭe ke saṃketakoṃ ke niyaṃtraṇa meṃ yugāṃtarakārī lābha anaṃtima khātoṃ (pīe) meṃ spaṣṭa hai |
| DictDis | 2013-14 meṃ **mukhya ghāṭā saṃketaka** kī rokathāma meṃ ullekhanīya ghāṭā anaṃtima khātoṃ (pīe) meṃ spaṣṭa hai |
| Reference | 2013-14 meṃ mukhya ghāṭā saṃketakoṃ para niyatramṇa se prāpta lābha ke bāre meṃ anaṃtima lekhe (pīe) meṃ spaṣṭa ullekha kiyā gayā hai |
| Input | Notwithstanding deterioration in export performance brought on, inter alia, by weak external market conditions, the current account deficit narrowed in 2014-15 from its level a year ago on terms of trade gains and weak import demand. |
| Constraints | ('export', 'niryāta')('inter alia', 'anya bātoṃ ke sātha-sātha') ('deterioration', 'kṣaya, kamī, girāvaṭa ')('terms of trade', 'vyāpāra kī śarteṃ') ('current account deficit', 'cālū khātā ghāṭā')('performance', 'pālana')('market', 'bājāra') |
| Transformer | **niryāta** pradarśana meṃ **girāvaṭa** ke bāvajūda kamajora bāharī bājāra paristhitiyoṃ ke kāraṇa **cālū khātā ghāṭā** eka sāla pahale ke stara se kama hokara eka sāla pahale 2014-15 meṃ vyāpāra lābha aura kamajora <u>āyāta māṃga</u> ke kāraṇa kama ho gayā thā |
| DictDis | **anya bātoṃ ke sātha-sātha** kamajora bāhya bājāra sthitiyoṃ ke kāraṇa **niryāta** ke pradarśana meṃ **girāvaṭa** ke bāvajūda cālū khātā ghāṭā varṣa 2014-15 ke apane stara se eka varṣa pahale vyāpāra lābha aura kamajora **āyāta śarteṃ** ke saṃdarbha meṃ kama ho gayā |
| Reference | anya bātoṃ ke sātha- sātha bāharī bājāroṃ kī kamajora hālata ke kāraṇa niryāta-niṣpādana meṃ āī girāvaṭa ke bāvajūda varṣa 2014-15 kā cālū khātā ghāṭā pichale varṣa ke isake stara se kama thā kyoṃki vyāpāra kī śartoṃ tathā āyāta kī kamajora māṃga se phāyadā huā thā |

Table 10: Phrasal constraint examples from Regulatory dataset using DICTDIS and Transformer (unconstrained). Here, the correctly ingested constraints are highlighted in **bold** and incorrectly ingested constraints are highlighted as <u>underline</u>.